\documentclass[runningheads]{llncs}
\usepackage{graphicx}

\usepackage{tikz}
\usepackage{comment}
\usepackage{amsmath,amssymb} 
\usepackage{color}
\usepackage{amssymb}
\usepackage{url}
\usepackage{graphicx}
\usepackage{wrapfig}
\usepackage{booktabs}
\usepackage{makecell}
\usepackage{colortbl}
\usepackage{multirow}
\usepackage{fixltx2e}
\usepackage{subcaption}
\usepackage{amsmath}
\usepackage{etoolbox}
\usepackage{multicol}
\usepackage{refcount}
\usepackage{hyperref}
\usepackage{marvosym}
\usepackage{ifsym}

\usepackage[accsupp]{axessibility}  
\definecolor{LightCyan}{rgb}{0.88,1,1}

\definecolor{LightCyan}{rgb}{0.88,1,1}
\definecolor{mygray}{gray}{0.9}
\definecolor{mygray2}{gray}{0.6}
\hypersetup{
    colorlinks=true,
    linkcolor=blue,
    filecolor=magenta,      
    urlcolor=cyan,
    pdftitle={Overleaf Example},
    pdfpagemode=FullScreen,
    }

\begin{document}
\pagestyle{headings}
\mainmatter
\def\ECCVSubNumber{4429}  

\title{UniNet: Unified Architecture Search with Convolution, Transformer, and MLP} 

\titlerunning{UniNet}
%
\author{Jihao Liu\inst{1,2} \and Xin Huang\inst{1} \and Guanglu Song\inst{2} \and Hongsheng Li\inst{1}\textsuperscript{\Letter} \and Yu Liu\inst{2}\textsuperscript{\Letter}}

\footnotetext{\textsuperscript{\Letter} Corresponding author.}
\authorrunning{Liu et al.}
%
\institute{CUHK, MMLab \and SenseTime Research}
\maketitle

\begin{abstract}
Recently, transformer and multi-layer perceptron (MLP) architectures have achieved impressive results on various vision tasks.
However, how to effectively combine those operators to form high-performance hybrid visual architectures still remains a challenge. In this work, we study the learnable combination of convolution, transformer, and MLP by proposing a novel unified architecture search approach. Our approach contains two key designs to achieve the search for high-performance networks.
First, we model the very different searchable operators in a unified form, and thus enable the operators to be characterized with the same set of configuration parameters. In this way, the overall search space size is significantly reduced, and the total search cost becomes affordable. Second, we propose context-aware downsampling modules (DSMs) to mitigate the gap between the different types of operators. Our proposed DSMs are able to better adapt features from different types of operators, which is important for identifying high-performance hybrid architectures. 
Finally, we integrate configurable operators and DSMs into a unified search space and search with a Reinforcement Learning-based search algorithm to fully explore the optimal combination of the operators. To this end, we search a baseline network and scale it up to obtain a family of models, named UniNets, which achieve much better accuracy and efficiency than previous ConvNets and Transformers. In particular, our UniNet-B5 achieves 84.9\% top-1 accuracy on ImageNet, outperforming EfficientNet-B7 and BoTNet-T7 with
44\% and 55\% fewer FLOPs respectively. By pretraining on the ImageNet-21K, our UniNet-B6 achieves 87.4\%, outperforming Swin-L with 51\% fewer FLOPs and 41\% fewer parameters.  Code is available at \url{https://github.com/Sense-X/UniNet}.





\end{abstract}

\section{Introduction}
\label{sec:intro}
Convolutional Neural Networks (CNNs) dominate the learning of visual representations and show effectiveness on various visual tasks, including image classification, object detection, semantic segmentation, etc. 
Recently, convolution-free backbones show impressive performances on image classification \cite{imagenet}. Vision Transformer (ViT) \cite{vit} demonstrates that pure transformer architecture that is mainly built on multi-head self-attentions (MSAs) can attain state-of-the-art performance when trained on large-scale datasets (e.g., ImageNet-21K, JFT-300M). 
MLP-Mixer \cite{mixer} introduced a pure multi-layer perceptron (MLP) architecture that can almost match ViT's performance without using the time-consuming attention mechanism. 
The main operators in those networks perform differently in terms of efficiency and data utilization.
On the one hand, convolutions in CNNs are locally connected and their weights are input-independent, which makes it effective at extracting low-level representations and efficient under the low-data regime. On the other hand, MSAs in the transformer capture long-range dependency, and the attention weights are dynamically dependent on the input representations. Hence, it is more data and computation demanding. The token-mixing in MLP-Mixer performs like a depthwise convolution of a full receptive field with parameter sharing, which is also data demanding. It is an important topic to study how to combine them effectively to form high-performance hybrid visual architectures, which, however, remains a challenge.

There were recent papers on attempting to manually combine the different types of operators to form hybrid visual networks. In ViT \cite{vit}, a hybrid architecture using ResNet and transformer is also studied and improves upon pure transformers for smaller model sizes. Besides, many other works \cite{convit,coatnet,yuan2021incorporating,cvt,cmt,nasvit,container} also explored the combination of convolution and transformer to form hybrid architectures to improve data or computation efficiency. Furthermore, the combination of convolution and MLP is studied in \cite{convmlp}, and the combination of gated MLP and MSA is studied in \cite{gmlp}. Those previous approaches focus on combining two distinct operators and can achieve satisfactory performances to some extent. However, a unified view and a systematical study are missed in prior arts.

We identify two key challenges when building high-performance hybrid architectures: (1) The operators can be implemented with various styles, and it is infeasible to manually explore all possible implementations and combinations. Although we can automate the exploration with Neural Architecture Search (NAS) techniques, the search space should be properly designed so that the search cost is affordable. (2) Each operator has its own characteristics, and simply combining them together does not lead to optimal results. We conduct a simple pilot study on directly stacking different operators to form hybrid networks. As shown in Table~\ref{pilot_study}, however, the straightforward stacking of different operators achieves even worse performance than the vanilla ViT.

In this paper, we study the learnable combination of convolution, transformer, and MLP by proposing a novel unified architecture search approach. Our approach has two key designs to address the challenges mentioned above. 
First, we model distinct operators in a unified form, and use the same
set of searchable configuration parameters (i.e., \textit{OP type}, \textit{expansion}, \textit{channels}, etc) to characterize each of the different operators.
The unified design enables us to greatly reduce the overall search space, and as a result, the total search cost becomes affordable. Besides, we propose context-aware downsampling modules (DSMs) to harmonize the combination of different operators. The proposed DSMs can be instantiated into three types, i.e., Local-DSM (L-DSM), Local-Global-DSM (LG-DSM), and Global-DSM (G-DSM), aiming to better adapt the representations from one operator to another. 
Based on these designs, we build a unified search space consisting of a large family of different general operators (GOPs), DSMs, and network size, and jointly optimize model accuracy and FLOPs for identifying high-performance hybrid networks. We illustrate the search space and the backbone in Figure~\ref{fig:backbone}.


\begin{figure*}[t]
    \centering
    \begin{subfigure}[b]{0.31\textwidth}
        \includegraphics[width=\textwidth]{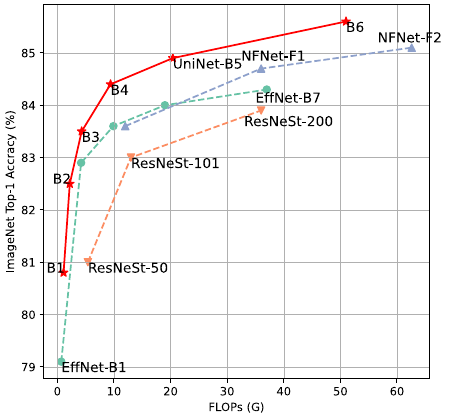}
        \caption{ConvNet}
    \end{subfigure}
    %
    \centering
    \begin{subfigure}[b]{0.31\textwidth}
        \includegraphics[width=0.99\textwidth]{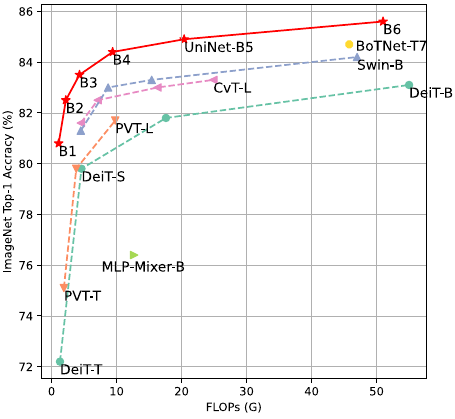}
        \caption{Transformer/Hybrid}
    \end{subfigure}
    \centering
    \begin{subfigure}[b]{0.31\textwidth}
        \includegraphics[width=0.99\textwidth]{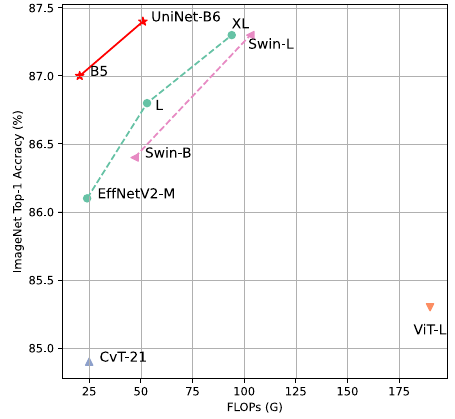}
        \caption{ImageNet21K Transfer}
    \end{subfigure}
    \caption{\textbf{ImageNet top-1 accuracy vs. FLOPs.} Our UniNet-B5 achieve 84.9\% with ImageNet-1K dataset, outperforming EfficientNet-B7 and BoTNet-T7 with 44\% and 55\% fewer FLOPs, respectively. Our UniNet-B6 achieve 87.4\% on ImageNet-1K with ImageNet-21K pre-training, outperforming EfficientNetV2-XL with 46\% fewer FLOPs.}
    \label{fig:sota}
\end{figure*}

\begin{table}[t]
\centering
\caption{ImageNet top-1 accuracy of different operator combinations. T, M, and C refer to transformer block, MLP-Mixer block, and Depthwise Convolution block respectively. Different block numbers are chosen so that their computations are comparable.}
\resizebox{0.8\linewidth}{!}{
    \begin{tabular}{l|ccc|c}
    \midrule
    Model & Configuration & \makecell{\#Params (M) } & \makecell{\#FLOPs (G)} & \makecell{Top-1 Acc.} \\ 
    \midrule
    ViT & 12 T & 22 & 4.6 & 78.0 \\ 
    MLP-Mixer & 18 M & 23 & 4.7 & 76.8 \\ 
    DWConv & 18 C & 22 & 4.3 & 78.1 \\ 
    ViT-MLP & 7 T + 7 M & 22 & 4.5  & 76.5 \\
    MLP-ViT   & 7 M + 7 T & 22 & 4.5 & 77.8 \\
    DWConv-ViT & 7 C + 7 T & 22 & 4.3 & 79.5 \\
    \bottomrule
    \end{tabular}
    }
\label{pilot_study}
\end{table}

The discovered network, named UniNet, exhibits strong performance and efficiency improvements over common ConvNets, Transformers, or hybrid architectures on various visual benchmarks. Our experiments show that UniNet has the following characteristics: (1) placing convolutions in the shallow layers and transformers in the deep layers, (2) allocating a similar amount of FLOPs for both convolutions and transformers, and (3) inserting L-DSM to downsample for convolutions and LG-DSM for transformers. Our analysis shows that the conclusion is consistent among the top-5 models.

To go even further, we build a family of high-performance UniNet models by scaling up the searched baseline network, which achieves better accuracy and efficiency in both small and large model sizes. In particular, our UniNet-B5 achieves comparable accuracy (+0.1\%) to EfficientNet-B7 while requires much less computation cost (-44\%) (Figure~\ref{fig:sota} (a)). By pretraining on large-scale ImageNet-21K, our UniNet-B6 achieves 87.4\% accuracy, outperforming Swin-L with fewer FLOPs (-51\%) and parameters (-41\%) (Figure~\ref{fig:sota} (c)).

\section{Related Works}
\label{related_works}

\noindent\textbf{Convolution, Transformer, and MLP.}
A host of ConvNets have been proposed to push forward the state-of-the-art computer vision approaches such as \cite{resnet,inception,efficientnet}. Despite the numerous CNN models, their basic operators, convolution, are the same. Recently, \cite{vit} proposed a pure transformer-based image classification model ViT, which achieves impressive performance on the ImageNet benchmark. DeiT \cite{deit} shows that well-trained ViT can obtain a better performance-speed trade-off than ConvNets. PVT \cite{pvt} and Swin \cite{swin} propose multi-stage vision transformers, which can be easily transferred to other downstream tasks. On the other hand, recent papers are attempting to use only MLP as the building block. MLP-Mixer \cite{mixer}, ResMLP \cite{resmlp}, and ViP \cite{vip} show that pure MLP architectures can also achieve near state-of-the-art performance.

\noindent\textbf{Combination of different operators.}
Another line of work tries to combine different operators to form new networks. CvT \cite{cvt} propose to incorporate self-attention and convolution by generating $\mathtt{Q}$, $\mathtt{K}$, and $\mathtt{V}$ in self-attention with convolution. CeiT \cite{yuan2021incorporating} replace the original patchy stem with a convolutional stem and add depthwise convolution to the FFN layer, which obtains fast convergence and better performance. ConViT \cite{convit} tries to unify convolution and self-attention with gated positional self-attention and is more sample-efficient than self-attention. Many other works \cite{coatnet,cmt,nasvit,container} also explored the combination of convolution and transformer to form hybrid architectures to improve the data or computation efficiency. Besides, ConvMLP \cite{convmlp} studied the combination of convolution and MLP, and gMLP \cite{gmlp} studied the combination of gated MLP and multi-head self-attentions (MSA). Instead of requiring manual exploration of the hybrid architectures, we propose a unified architecture search approach to automatically search for high-performance hybrid architecture.

\noindent\textbf{Downsampling module.}
In ConvNets, the downsampling module (DSM) is implemented with strided-Conv or pooling.  As DSM breaks the shift-invariant of convolution, \cite{zhang2019shiftinvar} propose anti-aliased DSM to keep it. Besides, a line of works tries to preserve more information when downsampling with a learnable or dynamic kernel \cite{lip,dpp,carafe++}. Most of their approaches are downsampling based on local context, which we show is not suitable for our unified network. In our work, we propose context-aware DSM and jointly search with operator combinations, which guarantees better performance.

\section{Method}
\label{method}

\begin{figure*}[t]
    \centering
    \includegraphics[width=0.85\textwidth]{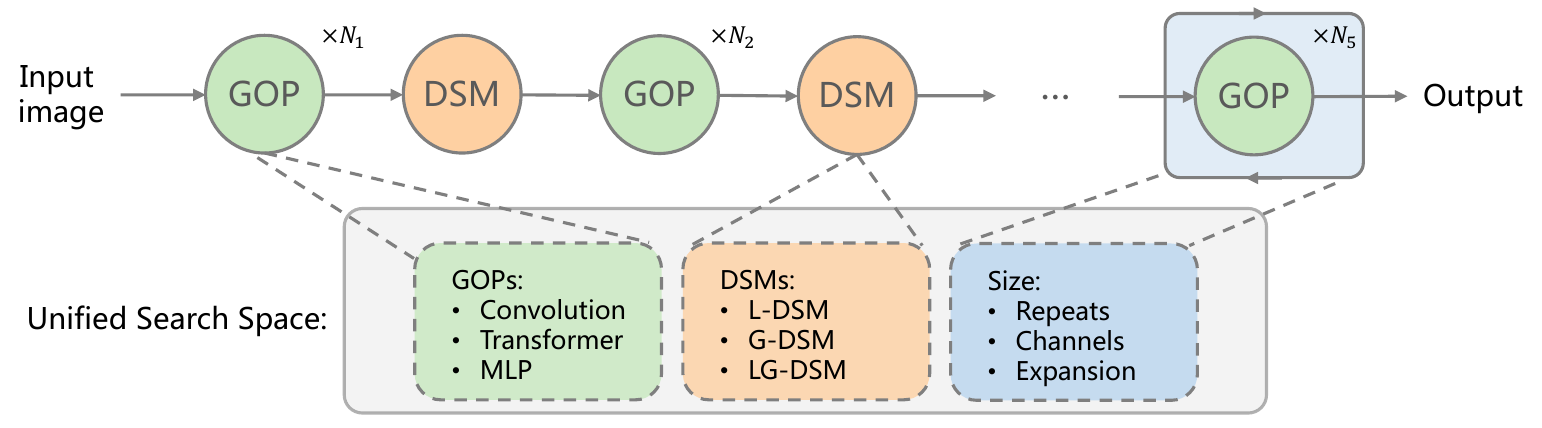}
    \caption{\textbf{Unified Architecture Search.} We jointly search different types of operators as well as downsampling modules (DSM) and network size in a unified search space. We construct UniNet architecture in a multi-stage fashion. Between two successive stages, one of the DSMs is inserted to change the spatial dimension or channels.}
    \label{fig:backbone}
\end{figure*}

\subsection{Unified Architecture Search}
\label{sec:uas}

As discussed in previous works \cite{convit}, an appropriate combination of convolution and transformer operators can lead to performance improvements. 
However, the previous approaches \cite{cvt,yuan2021incorporating} only adopt convolution in self-attention or feed-forward network (FFN) sub-layers and stack them repeatedly. Their approaches did not fully explore the combinations to take advantage of their different characteristics.

Prior arts \cite{carafe++,zhang2019shiftinvar} show that the downsampling module plays an important role in visual tasks. Most previous approaches adopt hand-crafted downsampling operations, i.e., strided convolution, max-pooling, or avg-pooling, to downsample the feature map based on only the local context. However, these operations are specifically designed for ConvNets, and might not be suitable to the transformer or MLP based architectures, which capture representation globally. 

In this paper, we investigate the learnable combination of convolution, transformer, and MLP\footnote{Here, MLP refers to a MLP-style sub-layer that captures spatial representations \cite{mixer,resmlp,vip}, instead of pure $1\times 1$ convolution.}, trying to assemble them to create high-performance hybrid visual network architectures. For better transmitting features across different operator blocks, we proposed context-aware downsampling modules. We jointly search the operators, downsampling modules, and network size in a unified search space. In contrast, previous Neural Architecture Search (NAS) works achieved state-of-the-art performances mainly via searching the network sizes. We show that the searched hybrid architecture by our unified architecture search approach can achieve very promising performance.

In the remaining parts of the section, we firstly present how to properly define different operators into a unified search space and search them jointly. We then present the challenge of incorporating downsampling modules with different operators and present our proposed context-aware downsampling module. Finally, we will introduce our UniNet architectures and NAS pipeline.

\subsection{Modeling Convolution, Transformer, MLP with a Unified Searchable Form}
\label{sec3.2}

Recently, transformer and MLP based architectures are able to achieve comparable performance to convolution networks on different visual tasks. To achieve better performance, it is intuitive to assemble all the types of operators to build high-performance hybrid networks.
Actually, a few works \cite{cvt,yuan2021incorporating,convit} have been studied to empirically combine convolution and self-attention. However, manually searching network architectures is quite time-consuming and cannot ensure optimal performances with different computational budgets.

We introduce a unified search space that contains General Operators (GOPs, including convolution, transformer, and MLP), and then search for the optimal combination of those operators jointly. Compared with the prior art, we propose a unified form to characterize different operators. Specifically, we use the inverted residual \cite{mbv2} to model a general operator block, which first expands the input channel $\mathtt{c}$ to a larger size $\mathtt{ec}$, and then projects the $\mathtt{ec}$ channels back to $\mathtt{c}$ for residual connection. The $\mathtt{e}$ is defined as the expansion ratio, which is usually a small integer number, e.g., 4. The general operation block is therefore modeled as
\begin{equation}
    \mathtt{y} = \mathtt{x + Operation(x)},
\end{equation}
where $\mathtt{Operation}$ can be convolution, MLP, or transformer, and $\mathtt{x, y}$ represent input and output features, respectively. For convolution, we place the convolution operation inside the bottleneck \cite{mbv2}, which can be expressed as
\begin{equation}
    \mathtt{Operation(x) = Proj_{ec\rightarrow c}(Conv(Proj_{c\rightarrow ec}(x))).}
\end{equation}
The $\mathtt{Conv}$ operation can be either regular convolution or depth-wise convolution ($\mathtt{DWConv}$) \cite{xception}, and the $\mathtt{Proj}$ represents a linear projection.
For self-attention in transformer and token-mixing in MLP, the computation cost on the large bottleneck feature map is quite huge. Following previous works \cite{vit,mixer}, we separate them from the bottleneck for computation efficiency, and the $\mathtt{Proj}$ is implemented inside the FFN \cite{attention} sub-layer. Each transformer block has a query-key-value self-attention sub-layer and an FFN sub-layer, and the token-mixing in the MLP block is implemented by transpose-FFN-transpose 
as that in \cite{mixer},
\begin{gather}
    \mathtt{y = y' + FFN(y')}, \\
    \mathtt{y' = x + SA(x) \; \textrm{or} \; x + MLP(x)}, \\
    \mathtt{FFN(y') = Proj_{ec \rightarrow c}(Proj_{c\rightarrow ec}(y')),}
\end{gather}
where $\mathtt{SA}$ can be either vanilla self-attention or local self-attention $\mathtt{LSA}$, and $\mathtt{MLP}$ refers to the token-mixing operation.

There are two main advantages of representing the different types of operators in a unified format and search space: (1) We can characterize each operator with the same set of configuration parameters (i.e., \textit{OP type}, \textit{expansion}, \textit{channels}, etc). As a result, the overall search space is greatly reduced, and the total search cost becomes affordable. (2) With the unified form, the comparison between different operators is fairer, which is important for NAS \cite{mnas} to identify the optimal hybrid architecture.


\subsection{Context-Aware Downsampling Modules}
\label{sec:dsm}

As discussed in Section \ref{sec:uas}, the downsampling module (DSM) plays an important role in visual tasks. In addition to hand-crafted DSM (i.e., max-pooling or avg-pooling), a few works \cite{dpp,lip,carafe++} tried to preserve more information via downsampling with the learnable or dynamic kernel. Most of the approaches utilized downsampling based on local context, which suits conventional ConvNets well. However, in our unified search space, operators with different receptive fields can be assembled unrestrictedly to form a hybrid architecture, where the local context might be destroyed and therefore the previous downsampling operations might not be suitable.

\begin{figure}[t]
    \centering
     \begin{subfigure}[b]{0.17\linewidth}
         \centering
         \includegraphics[width=\textwidth]{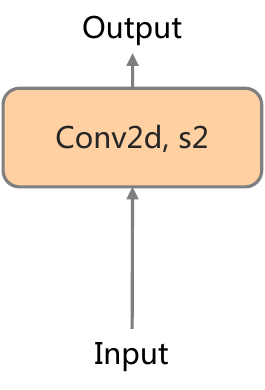}
         \caption{L-DSM}
         \label{fig:ldsm}
     \end{subfigure}
     \hspace{1.5em} 
     \begin{subfigure}[b]{0.17\linewidth}
         \centering
         \includegraphics[width=\textwidth]{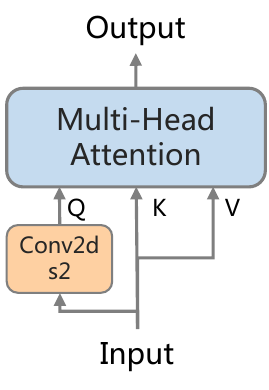}
         \caption{LG-DSM}
         \label{fig:lgdsm}
     \end{subfigure}
     \hspace{1.5em} 
     \begin{subfigure}[b]{0.17\linewidth}
         \centering
         \includegraphics[width=\textwidth]{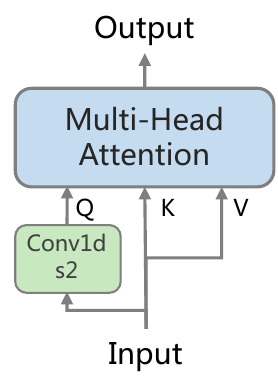}
         \caption{G-DSM}
         \label{fig:gdsm}
     \end{subfigure}
    \caption{Structures of the context-aware downsampling modules. The three DSMs are described in Section \ref{sec:dsm}. Shortcuts are omitted for better visualization.}
    \label{fig:dsms}
\end{figure}

In this paper, we propose context-aware DSM, which is instanced with Local-DSM (L-DSM), Local-Global-DSM (LG-DSM), and Global-DSM (G-DSM). The main difference between those DSMs is the considered context when performing downsampling. For L-DSM, only local context is involved, which fits ConvNets well as shown in previous works \cite{pvt,swin}. For G-DSM, only global context is used for downsampling, which may fit other operators, e.g., transformers. The LG-DSM combines the characteristics of L-DSM and G-DSM. It uses both local and global context for downsampling. Our intuition is that one of the largest dissimilarities of different operators is the receptive field. Transformer and MLP naturally have global receptive filed, while convolution has local receptive field, e.g., $3 \times 3$. When combining those operators, there is no single optimal DSM that satisfies all scenarios.

The proposed DSMs are visualized in Figure \ref{fig:dsms}. To downsample based on global cues, we utilize the self-attention mechanism to capture global context, which is missed by the prior art. 
For G-DSM, we use $\mathtt{Conv1D}$ with stride 2 to downsample the query and use the downsampled query features to aggregate key features with downsampled output resolution.
Note that, there is no local context preserved after downsampling of G-DSM. For LG-DSM, we first reshape the flattened token sequences back to the spatial grid and apply $\mathtt{Conv2D}$ with stride 2 to downsample the query, and then flatten the query back to calculate the attention weights.

Compared with previous works, which mainly try to improve ConvNets, our proposed DSMs are not designed for a specific architecture. Our motivation is that different DSMs might be suitable for different operators. For example, the optimal DSM might be L-DSM for ConvNets, but G-DSM for transformers. As thousands of operator combinations would be trained in our NAS process, it is unfeasible to decide which DSM to use by hand. To obtain the optimal architecture, we jointly search DSMs with other operators. 
In our searched optimal architecture, L-DSM is indeed used between operators with the local receptive field while LG-DSM is favored by operators with a global receptive field. The results validate the effectiveness of our proposed context-aware downsampling modules.

\subsection{UniNet Architecture}

As shown in recent studies, combining different operators \cite{cvt,yuan2021incorporating} can bring performance improvements. Most previous approaches only repeatedly stack the same operator in the whole architecture and search only different channels in different stages. These approaches do not allow large architecture diversity in each block, which we show is crucial for achieving high accuracy for hybrid architectures.

On the contrary, in our UniNet, the operators are not fixed but searched from the unified search space. We construct our UniNet architecture in a multi-stage fashion, which can be easily transferred to downstream tasks. Between two successive stages, one of our proposed DSMs is inserted to reduce the spatial dimension. We jointly search the GOP and DSM for all stages. The GOP could be different for different stages but repeated multiple times in one stage, which can greatly reduce the search space size as pointed out before \cite{mnas}. The overall architecture and unified search space are illustrated in Figure \ref{fig:backbone}. 

Thanks to our unified form of GOPs, the network size of each stage can be configured with the repeat number $\mathtt{r}$, channel size $\mathtt{c}$, and expansion ratio $\mathtt{e}$. To obtain better computation-accuracy trade-off, we jointly search the network size with the GOP and DSM. For GOP, we search for convolution, transformer, MLP, and their promising variants, i.e., \{$\mathtt{SA}$, $\mathtt{LSA}$, $\mathtt{Conv}$, $\mathtt{DWConv}$, $\mathtt{MLP}$\}, 
as defined in Section \ref{sec3.2}; for $\mathtt{e}$, we search from \{2, 3, 4, 5, 6\}. The $\mathtt{LSA}$ refers to the window self-attention with window size of $7\times7$. Please note that we do not use shifted window like Swin Transformer~\cite{swin}.
The kernel size for convolution operation is fixed to 3$\times$3. The head dimension in self-attention is fixed to 32. 
We start the architecture search with an initial architecture, whose network size is determined based on a reference architecture, e.g., EfficientNetV2 \cite{effnetv2}. The initial channels and repeats are set according to the reference architecture.
For $\mathtt{c}$ and $\mathtt{r}$, we search from the sets \{0.5, 0.75, 1.0, 1.25, 1.5\} and \{-2, -1, 0, 1, 2\}, respectively. Channels are set to be divisible by 32 for self-attention.
Suppose we partition the network into $K$ stages, and each stage has a sub-search space of size $S$. Then the total search space is $S^K$. In our implementation, $K$ is set to 5 and $S$ equals 1,875. As a result, our search space size is about 2$\times$10\textsuperscript{16} and covers a large set of operators with quite different characteristics.

\subsection{Search Algorithm}

We use Reinforcement Learning (RL)-based search algorithm to search for high-performance hybrid architecture in our unified search space by jointly optimizing the model accuracy and FLOPs. Concretely, we follow previous work \cite{fnas,mnas} and map an architecture in the unified search space to a list of tokens, which are determined by a sequence of actions generated by a Recurrent Neural Network (RNN). 
The RNN is optimized with the PPO algorithm \cite{ppo} by maximizing the expected reward. In our implementation, we simultaneously optimize accuracy and the theoretical computation cost (FLOPs). To handle the multi-objective optimization problem, we use a weighted product customized as \cite{mnas} to approximate Pareto optimal. For a sampled architecture $m$, the reward is formulated as $r(m) = a(m) \times (\frac{t}{f(m)})^{\alpha}$, where function $a(m)$ and $f(m)$ return the accuracy and the FLOPs of $m$, $t$ is the target FLOPs, and $\alpha$ is a weight factor that balances the accuracy and computation cost. We include more details of the RL algorithm in the supplementary materials.

During the search process, thousands of combinations of GOPs and DSMs are trained on a proxy task with the same setting, which gives us a fair comparison between those combinations.  When the search is over, the top-5 architectures with the highest reward are trained with full epochs, and the top-performing one is kept for model scaling and transferring to other downstream tasks.

\section{Experimental Setup and Implementation}
To find the optimal architecture in our search space, we directly search on the large-scale dataset, ImageNet-1K. We reserve 50k images from the training set as a validation set. We employ a proxy task setting in the search phase. For each sampled architecture, we train it for 5 epochs and calculate the reward of the architecture with its FLOPs and the accuracy on the validation set. We set the target FLOPs $t$ and weight factor $\alpha$ in the reward function to 550M and 0.07 respectively~\cite{efficientnet}. During the search process, totally 2K models are trained on the proxy task. After that, we fully train the top-5 architectures on ImageNet-1K and preserve the top-performing one for model scaling and transferring to other downstream tasks.

For full training on the ImageNet-1K dataset, we follow the popular training recipe in DeiT \cite{deit}. We employ AdamW optimizer \cite{adam} with an initial learning rate of 0.001 and weight decay of 0.05 to train UniNet. The total batch size is set to 1024. We totally train for 300 epochs with a cosine learning rate decay and 5 epochs of linear warm-up. We follow the augmentation strategy in DeiT \cite{deit} and apply small augmentation for small models and heavy augmentation for large models as introduced in \cite{cait,trainvit}.
For training efficiency, UniNet-B5 and UniNet-B6 are trained with $224 \times 224$ input size and then finetuned on the large resolution. 
We also pre-train UniNet on a larger ImageNet-21K dataset, which contains 14.2 million images and 21K classes, to further test UniNet. We pretrain for 90 epochs with AdamW optimizer. We then finetune on ImageNet-1K for 30 epochs and compare the top-1 accuracy on ImageNet-1K with other approaches.
We list the details of training and finetuning hyper-parameters in the supplementary materials.

Besides, we also transfer UniNet to downstream tasks, e.g., object detection and instance segmentation on COCO and semantic segmentation on ADE20K.
For COCO training, we use the various detection frameworks and train UniNet with the widely-used 1x (12 epochs) and 3x (36 epochs) schedules. For ADE20K training, we use the UperNet framework and train with the same setting as \cite{swin}. The training details are listed in the supplementary materials.

\begin{table}[t]
    \begin{minipage}[t]{0.56\linewidth}
        \centering
        \caption{UniNet-B0 architecture. GOP and DSM represent General Operators and downsampling module respectively. $\mathtt{DWConv}$ and $\mathtt{SA}$ are described in Section \ref{sec3.2}.}
        \resizebox{0.85\linewidth}{!}{
        \begin{tabular}{c|cc|ccc|c}
        \toprule
        \multirow{2}{*}{Stage}  & \multicolumn{2}{c}{Operator} & \multicolumn{3}{c}{Network Size} & \multirow{2}{*}{FLOPs(M)} \\
         & GOP & DSM & $\mathtt{e}$ & $\mathtt{c}$ & $\mathtt{r}$ & \\
        \midrule
        0     & $\mathtt{DWConv}$ & L-DSM & 4 & 48    & 2 & 68 \\
        1     & $\mathtt{DWConv}$ & L-DSM & 6 & 80    & 4 & 135 \\
        2     & $\mathtt{DWConv}$ & L-DSM & 3 & 128   & 4 & 42 \\
        3     & $\mathtt{SA}$ & LG-DSM & 2 & 128   & 4 & 63 \\
        4     & $\mathtt{SA}$ & LG-DSM & 5 & 256   & 8  & 187 \\
        \bottomrule
        \end{tabular}
        }
        \label{tab:uninet_b05}
    \end{minipage}
    \hspace{1em}
    \begin{minipage}[t]{0.4\linewidth}
        \centering
        \caption{Performance of Top-5 models after fully training. D and A are short for $\mathtt{DWConv}$ and $\mathtt{SA}$ respectively.}
        \resizebox{0.65\linewidth}{!}{
        \begin{tabular}{ccc}
            \toprule
            Rank & Configuration & \makecell{Top-1 \\ Acc.} \\
            \midrule
            0 & \textbf{DDDAA} & \textbf{79.1} \\
            1 & DDDAA & 78.7 \\
            2 & DDDAD & 77.9 \\
            3 & DDDAA & 78.6 \\
            4 & DDDAA & 78.4 \\
            \bottomrule
        \end{tabular}
        }
        \label{tab:top5}
    \end{minipage}
\end{table}

\begin{table}[t]
    \begin{minipage}[t]{0.48\linewidth}
        \caption{UniNet performance on ImageNet. All UniNet models are trained on the ImageNet-1K dataset with 1.28M images. C, T, and H denote convolution, transformer, and hybrid architecture respectively.}
        \centering
        \resizebox{0.95\textwidth}{!}{
            \begin{tabular}{lc|ccc|c}
            \toprule
            \multirow{2}{*}{Model} & \multirow{2}{*}{Family} & \multirow{2}{*}{\makecell{Input \\ Size}} & \multirow{2}{*}{\makecell{\#FLOPs \\ (G)} } & \multirow{2}{*}{\makecell{\#Params \\ (M)}} & \multirow{2}{*}{\makecell{Top-1 \\ Acc.}} \\
            & & & & & \\
            \midrule
            EffNet-B0 \cite{efficientnet} & C     & 224   & 0.39  & 5.3   & 77.1 \\
            EffNetV2-B0 \cite{effnetv2} & C     & 240   & 0.7   & 7.4   & 78.7 \\
            DeiT-Tiny \cite{deit} & T     & 224   & 1.3   & 5.7   & 72.2 \\
            PVT-Tiny \cite{pvt} & T     & 224   & 1.9   & 13.2  & 75.1 \\
            ConViT-Ti+ \cite{convit} & H     & 224   & 2     & 10    & 76.7 \\
            \rowcolor{LightCyan}
            UniNet-B0 & H     & 160   & 0.56  & 11.5  & 79.1 \\
            \midrule
            EffNet-B2 \cite{efficientnet} & C     & 260   & 1     & 9.2   & 80.1 \\
            EffNetV2-B1 \cite{effnetv2} & C     & 260   & 1.2   & 8.1   & 79.8 \\
            RegNetY-4G \cite{regnet} & C     & 224   & 4     & 20.6  & 81.9 \\
            DeiT-Small \cite{deit} & T     & 224   & 4.3   & 22    & 79.8 \\
            PVT-Small \cite{pvt} & T     & 224   & 3.8   & 24.5  & 79.8 \\
            \rowcolor{LightCyan}
            UniNet-B1 & H     & 224   & 1.1  & 11.5    & 80.8 \\
            \midrule
            EffNet-B3 \cite{efficientnet} & C     & 300   & 1.8   & 12    & 81.6 \\
            EffNetV2-B3 \cite{effnetv2} & C     & 300   & 3     & 14    & 82.1 \\
            Swin-T \cite{swin} & T     & 224   & 4.5   & 29    & 81.3 \\
            CoAtNet-0 \cite{coatnet} & H & 224 & 4.2 & 25 & 81.6 \\
            \rowcolor{LightCyan}
            UniNet-B2 & H     & 256   & 2.2   & 16.2 & 82.5  \\
            \midrule
            EffNet-B4 \cite{efficientnet} & C     & 380   & 4.2   & 19    & 82.9 \\
            NFNet-F0 \cite{nfnet} & C     & 256   & 12.4  & 71.5  & 83.6 \\
            Swin-B \cite{swin} & T     & 224   & 15.4  & 88    & 83.5 \\
            ConViT-B+ \cite{convit} & H     & 224   & 30    & 152   & 82.5 \\
            CoAtNet-1 \cite{coatnet} & H & 224 & 8.4 & 42 & 83.3 \\
            CvT-21 \cite{cvt} & H     & 384   & 24.9  & 32    & 83.3 \\
            \rowcolor{LightCyan}
            UniNet-B3 & H     & 288   & 4.3   & 24  & 83.5 \\
            \midrule
            EffNet-B7 \cite{efficientnet} & C     & 600   & 37    & 66    & 84.3 \\
            EffNetV2-M \cite{effnetv2} & C     & 480   & 24    & 54    & 85.1 \\
            NFNet-F2 \cite{nfnet} & C     & 352   & 62.6  & 193.8 & 85.1 \\
            BoTNet-T7 \cite{botnet} & T     & 384   & 45.8  & 75.1  & 84.7 \\
            CoAtNet-1 \cite{coatnet} & H & 384 & 27.4 & 42 & 85.1 \\
            \rowcolor{LightCyan}
            UniNet-B4 & H     & 320   & 9.4   & 43.8  & 84.4 \\
            \rowcolor{LightCyan}
            UniNet-B5 & H     & 384   & 20.4  & 72.9  & 84.9 \\
            \rowcolor{LightCyan}
            UniNet-B6 & H     & 448   & 51  & 117  & 85.6 \\
            \bottomrule
            \end{tabular}%
        }
        \label{tab:sota}%
    \end{minipage}
    \hfill
    \begin{minipage}[t]{0.48\linewidth}
        \centering
        \caption{Performance on ImageNet with ImageNet-21K pre-train. All models are pre-trained on ImageNet-21K and finetuned on ImageNet-1K.}
        \resizebox{0.95\linewidth}{!}{
            \begin{tabular}{lc|ccc|c}
            \toprule
            Model & Family & \makecell{Input \\ Size} & \makecell{\#FLOPs \\ (G)}& \makecell{\#Params \\ (M)} & \makecell{Top-1 \\ Acc.} \\
            \midrule
            EffNetV2-M \cite{effnetv2} & C & 480 & 24 & 55 & 86.1 \\
            ViT-L/16 \cite{vit} & T & 384 & 190.7 & 304 & 85.3 \\
            HaloNet-H4 \cite{halonet} & T & 384 & - & 85 & 85.6 \\
            Swin-B \cite{swin} & T & 384 & 47.1 & 88 & 86.4 \\
            CvT-21 \cite{cvt} & H & 384 & 25 & 32 & 84.9 \\
            \rowcolor{LightCyan}
            UniNet-B5 & H & 384 & 20.4 & 72.9 & 87 \\
            \midrule
            EffNetV2-L \cite{effnetv2} & C & 480 & 53 & 121 & 86.8 \\
            EffNetV2-XL \cite{effnetv2} & C & 512 & 94 & 208 & 87.3 \\
            Swin-L \cite{swin} & T & 384 & 103.9 & 197 & 87.3 \\
            CoAtNet-2 \cite{coatnet} & H & 384 & 49.8 & 75 & 87.1 \\
            CoAtNet-2 \cite{coatnet} & H & 512 & 96.7 & 75 & 87.3 \\
            \rowcolor{LightCyan}
            UniNet-B6 & H & 448 & 51 & 117 & 87.4 \\
            \bottomrule
            \end{tabular}
        }
        \label{tab:21k}
        \centering
        \caption{Comparison with previous efficient architectures. UniNet is trained with knowledge distillation for a more fair comparison.}
        \resizebox{0.9\linewidth}{!}{
            \begin{tabular}{lc|cc}
            \toprule
            Model & Family & \makecell{\#FLOPs (M)} & \makecell{Top-1 Acc.} \\
            \midrule
            AttentiveNAS \cite{attentivenas} & C & 491 & 80.1 \\
            AlphaNet \cite{alphanet} & C & 491 & 80.3 \\
            FBNetv3 \cite{fbnetv3} & C & 557 & 80.5 \\
            OFA \cite{ofa} & C & 595 & 80.0 \\
            LeViT \cite{levit} & H & 658 & 80.0 \\
            \rowcolor{LightCyan}
            UniNet-B0 & H & 555 & 80.8 \\
            \bottomrule
            \end{tabular}
        }
        \label{tab:nas}
    \end{minipage}
    
\end{table}

\section{Main Results}
\label{main_results}

In this section, we firstly present our searched UniNet architecture. We then show the performance of the scaled UniNets on classification, object detection, and semantic segmentation.

\subsection{UniNet Model Family}
Table \ref{tab:uninet_b05} shows our searched UniNet-B0 architecture. Our searched architecture has the following characteristics: (1) Placing convolution in the shallow layers and transformers with $\mathtt{SA}$ in the deep layers. While the previous work \cite{vit} shows that the early-stage transformer blocks learn to gather local representations, our searched architecture directly applies convolution at early stages, which is more efficient. We further compare the top-5 searched models in Table \ref{tab:top5}, and find the conclusion is close to consistent. The exception is the 3rd model, which uses $\mathtt{DWConv}$ at the last stage, but with inferior performance. 
(2) Allocating a similar amount of computations for
both convolutions and transformers. Shown in Table \ref{tab:uninet_b05}, the $\mathtt{DWConv}$ stages consume 245M FLOPs, and $\mathtt{SA}$ stages consume 250M FLOPs. While the operator combination has been studied in prior arts, the computation allocating for different operators is neglected. Our work shed some light on this question by jointly searching the network size in our unified search space.
(3) Inserting L-DSM to downsample for convolutions and LG-DSM for transformers. Our search results show that the widely-used downsampling module is sub-optimal for hybrid architectures. We also notice that the MLP operator has not been chosen in the searched UniNet. We empirically find that the MLP-style operation breaks the spatial structure which is important for visual tasks \cite{islam2020much}, leading to inferior performance when combined with other operators. We add the visualization in the supplementary materials. 

To go even further, we build a family of high-performance UniNet models by scaling up the searched UniNet-B0. We utilize the compound scaling \cite{efficientnet} to scale depth, width, and resolution simultaneously. Note that the resolution is scaled with a smaller coefficient compared to EfficientNet \cite{efficientnet} for training and memory efficiency. We list the details of UniNet-B1 to UniNet-B6 in the supplementary materials.
While most previous transformer-based architectures outperform convolution-based architectures in large model sizes but underperform in small model sizes, UniNet achieves consistently better accuracy and efficiency across B0 to B6.

\subsection{ImageNet Classification Performance}
\noindent\textbf{ImageNet-1K.} Table~\ref{tab:sota} presents the performance comparison of our searched UniNet with previous proposed architectures. Our searched UniNet has better accuracy and computation efficiency than previous ConvNets, Transformers, or hybrid architectures. 

As shown in Table \ref{tab:sota}, under mobile setting, our UniNet-B0 achieves 79.1\% top-1 accuracy with 555M FLOPs, outperforming EfficientNetV2-B0 \cite{effnetv2} with less FLOPs. In the middle FLOPs setting, our UniNet-B3 achieves 83.5\% top-1 accuracy with 4.3G FLOPs, which outperforms the pure convolution-based EfficientNet-B4, pure transformer-based Swin-B, and hybrid architecture CvT-21. For larger models, our UniNet-B5 achieves 84.9\% with 20G FLOPs, outperforming EfficientNet-B7 and BoTNet-T7 with 44\% and 55\% fewer FLOPs, respectively. Figure \ref{fig:sota} (a, b) further visualizes the comparison of UniNet with other architectures in terms of accuracy and FLOPs. 

We further compare UniNet-B0 to previous searched efficient architectures in Table \ref{tab:nas}. Note that for a more fair comparison, we train UniNet-B0 with knowledge distillation. The details of distillation are listed in the supplementary materials. Shown in Table \ref{tab:nas}, UniNet-B0 achieves 80.8\% accuracy with 555M FLOPs, outperforming other efficient convolution-based or hybrid architectures.

\noindent\textbf{ImageNet-21K.} Table~\ref{tab:21k} presents the performance comparison of UniNet and other architectures with ImageNet-21K pretrain. Notably, UniNet-B5 obtains 87\% top-1 accuracy, which outperforms Swin-L with 4$\times$ less computation. UniNet-B6 achieves 87.4\% top-1 accuracy, which outperforms CoAtNet-2 \cite{coatnet} with 47\% less computation. We further visualize the comparison in Figure \ref{fig:sota} (c). 

\begin{table*}[t]
 \centering
\caption{Object detection, instance segmentation, and semantic segmentation performance on the COCO val2017 and ADE20K val set. All UniNet models are pre-trained on the ImageNet-1K dataset.}
\resizebox{0.9\textwidth}{!}{
  \begin{tabular}{l|cc|cc|cc|cc}
\toprule
\multirow{2}{*}{Backbone} & \multirow{2}{*}{\makecell{\#Params (M) \\ Det/Seg}} & \multirow{2}{*}{\makecell{\#FLOPs (G) \\ Det/Seg}} & \multicolumn{2}{c}{Mask R-CNN 1x} & \multicolumn{2}{c}{Mask R-CNN 3x} & \multirow{2}{*}{\makecell{UperNet \\ mIoU (\%)}} \\  
 & & & AP@box & AP@mask & AP@box & AP@mask &  \\ 
 \midrule
 ResNet18 \cite{resnet}   & 31/ - & 207/885 & 34.0 & 31.2 & 36.9 & 33.6 & -  \\
 ResNet50 \cite{resnet}  & 44/ - & 260/951 & 38.0 & 34.4 & 41.0 & 37.1 & -  \\
 PVT-Tiny \cite{pvt}  & 33/ - & 208/945 & 36.7 & 35.1 & 39.8 & 37.4 & -  \\
 \rowcolor{LightCyan}
 UniNet-B1 & 28/38  & 211/877 & 40.5 & 37.5 & 44.4 & 40.1 & 42.7 \\
 \midrule
 ResNet101 \cite{resnet} & 63/86  & 336/1029 & 40.4 & 36.4 & 42.8 & 38.5 & 44.9  \\
 PVT-Small \cite{pvt} & 44/ -  & 245/1039 & 40.4 & 37.8 & 43.0 & 39.9 & -  \\
 Swin-T  \cite{swin}  & 48/60  & 267/945 & 43.7 & 39.8 & 46.0 & 41.6 & 44.5  \\ 
 \rowcolor{LightCyan}
 UniNet-B3 & 42/51 & 270/940 & 45.2 & 41.1 & 47.9 & 42.9 & 48.5 \\
 \bottomrule
\end{tabular}
}
 \label{tab:det_seg_results}
\end{table*}

\begin{table}[t]
    \centering
    \caption{Performance on the COCO val2017 with various detection frameworks. The AP@box is reported. }
    \resizebox{0.8\textwidth}{!}{
    \centering
    \begin{tabular}{lcccc}
        \toprule
        Framework & Cascade-Mask-R-CNN & ATSS & Sparse-R-CNN & Mask-R-CNN \\
        \midrule
        ResNet50 \cite{resnet} & 46.3 & 43.5 & 44.5 & 41.0 \\
        Swin-T \cite{swin} & 50.5 & 47.2 & 47.9 & 46.0 \\
        \rowcolor{LightCyan}
        UniNet-B3 & 51.3 & 49.8 & 48.9 & 47.9 \\
        \bottomrule
    \end{tabular}
    }
    \label{tab:det_framwork}
    
\end{table}

\subsection{Object Detection and Semantic Segmentation Performance}
For object detection and semantic segmentation, we pick UniNet-B1 and UniNet-B3 and use them as the backbone networks for detection and segmentation frameworks. We compare our UniNet with other convolution or transformer-based architectures. For COCO object detection, we use various detection frameworks and compare the performance under 1$\times$ and 3$\times$ schedules. For ADE20K semantic segmentation we use the UperNet framework and report mIoU (\%) for different architectures under the same training setting.

As shown in Table \ref{tab:det_seg_results}, our searched UniNet consistently outperforms convolution-based ResNet \cite{resnet} and transformer-based PVT \cite{pvt} or Swin-Transformer \cite{swin}. UniNet-B1 achieves 40.5 AP@box, which is 3.8\% better than PVT-Tiny but with 15\% fewer parameters. UniNet-B3 achieves 45.2 AP@box with 1$\times$ schedule and 47.9 AP@box with 3$\times$ schedule, which is 1.5\% and 1.9\% better than Swin-T, respectively. We further test various detection framework and show the results in Table \ref{tab:det_framwork}, and find that UniNet achieves consistently better performance among others. 
For ADE20K semantic segmentation, we achieve 48.5\% mIoU with 51M parameters. Compared with transformer-based Swin-T, our UniNet outperforms 4.0\% mIoU with a similar parameter size. Besides, compared with convolution-based ResNet101, we achieve 3.6\% higher mIoU with 41\% fewer parameters. All the results show the effectiveness of our searched UniNet.

\section{Ablative Studies and Analysis}
\label{ablation}

In this section, we study the impact of joint search of General Operators and discuss the importance of context-aware downsampling modules (DSMs).

\subsection{Single Operator vs. General Operators}
\begin{table}[t]
    \caption{Performance on ImageNet with different search settings. One type of operator is kept for comparison with the hybrid UniNet.}
    \centering
        \resizebox{0.6\linewidth}{!}{
        \begin{tabular}{c | cc | c}
        \midrule
        Model & \makecell{\#FLOPs (G)} & \makecell{\#Params (M) } & {Top-1 Acc.} \\
        \midrule
        UniNet-B0   & 0.56     & 11.5     & \textbf{79.1} \\
        \midrule
        Convolution-Only & 0.59     & 11.0     & 77.7 \\
        Transformer-Only & 1.2      & 11.2     & 78.2 \\
        MLP-Only         & 0.95     & 11.4     & 76.8 \\
        \midrule
        \end{tabular}
        }
    \label{tab:with_conv}
\end{table}

Previous works \cite{mnas,efficientnet} mostly focus on the network size search, which uses a single operator, convolution, as the main feature extractor. In comparison, we jointly search the combination of different General Operators (GOPs), i.e., convolution, transformer, MLP, and their promising variants. To verify the importance of GOPs, we keep only one type of operator in the search space and re-run the search experiments under the same settings. After the search, we fully train the top-5 architectures with the highest reward on ImageNet-1K and report the best performance. 

As shown in Table \ref{tab:with_conv}, our searched hybrid architecture consistently achieves better accuracy compared to single-operator-based architectures. The result verifies the effectiveness of our unified architecture search of GOPs, which can take advantage of the characteristics of different operators. 

\subsection{Fixed vs. Context-Aware downsampling}

When combining different operators into a unified network, the traditional downsampling module, such as strided-conv or pooling, could be sub-optimal. To verify the effectiveness of our proposed context-aware DSMs, we replace the DSMs of our search UniNet with one fixed DSM and compare their performance under the same training setting.

As shown in Table \ref{tab:dsm}, our searched UniNet consistently outperforms its variants that use a single-fixed DSM in all stages. Although we see that using G-DSM or LG-DSM in all stages brings more computation and parameters, the performance does not become better. The result emphasizes the importance of our joint search of GOPs and DSMs.

Besides, we transfer our proposed DSMs to other popular transformer-based architectures, Swin-Transformer \cite{swin} and PVT \cite{pvt}. Both Swin and PVT have 4 stages. We compare 2 settings: 1) using LG-DSM for 4 stages, as both PVT and Swin are pure transformer architectures 2) using L-DSM for the first two stages while LG-DSM for the latter two stages, which requires less computation. As shown in Table \ref{tab:transfer_dsm}, our proposed LG-DSM improves PVT-Tiny and Swin-T for 3.5\% and 0.7\%, respectively. Using L-DSM in the first two stages has a similar computation compared with the baseline, which improves PVT-Tiny and Swin-T for 2.4\% and 0.4\%, respectively.
To note that, PVT uses a strided-conv for downsampling. As discussed in Section \ref{sec:dsm}, it is harmful to the main operator in PVT, which has a global receptive field. On the contrary, our proposed DSMs are able to downsample based on both local and global context, and can greatly improve the performance.

We integrate the proposed DSM into other architectures. We adopt the setting in rows 3/6 of Table 11 (i.e., L$\rightarrow$LG-DSM) to avoid introducing much computation cost. As shown in Table~\ref{tab:dsm}, we achieve consistent performance gains.

\begin{table}[t]
    \begin{minipage}[t]{0.48\linewidth}
    \centering
    \caption{Performance on ImageNet of UniNet with different DSMs. Note that the traditional strided-conv downsampling module is shown in row 2.}
    \resizebox{1.0\linewidth}{!}{
    \begin{tabular}{l | cc | c}
        \toprule
        Model & \#FLOPs (G) & \#Params (M)  & Top-1 Acc. \\
        \midrule
        UniNet            & 0.56     & 11.5     & \textbf{79.1} \\
        w/ L-DSM   & 0.54     & 11.3     & 78.5 \\
        w/ G-DSM   & 0.77     & 12.7     & 76.8 \\
        w/ LG-DSM  & 0.72     & 14.1     & 78.9 \\
        \bottomrule
    \end{tabular}
    }
    \label{tab:dsm}
    \end{minipage}
    \hfill
    \begin{minipage}[t]{0.48\linewidth}
    \centering
    \caption{Performance comparison on ImageNet of different backbones when equipped with our proposed DSMs.}
    \resizebox{1.0\linewidth}{!}{
    \begin{tabular}{l|cc|c}
        \toprule
        Model & \#FLOPs (G) & \#Params (M)  & Top-1 Acc. \\
        \midrule
        PVT-Tiny  \cite{pvt} & 1.9 & 13.2 & 75.1 \\ 
        w/ LG-DSM        & 3.1 & 17.3 & 78.6 \\
        w/ L$\rightarrow$LG-DSM & 2.0 & 14.3 & 77.5 \\
        \midrule
        Swin-T \cite{swin} & 4.5 & 29.0 & 81.2 \\
        w/ LG-DSM     & 6.4 & 33.4 & 81.9 \\
        w/ L$\rightarrow$LG-DSM & 4.7 & 30.0 & 81.6 \\
        \bottomrule
    \end{tabular}
    }
    \label{tab:transfer_dsm}
    \end{minipage}
\end{table}

\begin{table}[h]
    \centering
    \caption{Applying proposed DSM to various architectures.}
    \resizebox{0.6\linewidth}{!}{
    \begin{tabular}{cccccc}
        \toprule
         & EffNet-B1 & Swin-S & Swin-B & PVT-S & PVT-M \\
        \midrule
        w/o our DSM & 79.1 & 83.2 & 83.5 & 79.8 & 81.2 \\
        \midrule
        w/ our DSM & \textbf{79.6} & \textbf{83.4} & \textbf{83.9} & \textbf{81.6} & \textbf{82.3} \\ 
        \bottomrule
    \end{tabular}
    }
    \label{tab:dsm}
\end{table}




\section{Conclusion}
\label{conclusion}

In this paper, we propose a novel unified architecture search approach to jointly search the combination of convolution, transformer, and MLP. We empirically identify that the widely-used downsampling modules become the performance bottlenecks when the operators are combined. To further improve the performance, we propose context-aware downsampling modules and jointly search them with all operators. We scale the search baseline network up and obtain a family of models, named UniNet, which achieve much better accuracy and efficiency than previous ConvNets and Transformers.


\textbf{Acknowledgement}
Hongsheng Li is also a Principal Investigator of Centre for Perceptual and Interactive Intelligence Limited (CPII). This work is supported in part by CPII, in part by the General Research Fund through the Research Grants Council of Hong Kong under Grants (Nos. 14204021, 14207319), in part by CUHK Strategic Fund.

\section{Appendix}

\subsection{UniNet Implementations}
To reduce the search space size, we do not search for the kernel size of convolution operators. The kernel size is fixed to 3$\times$3. The head dimension in self-attention (i.e., in transformer block or LG-DSM) is fixed to 32. Following prior arts \cite{resnet,vit}, we use BN \cite{bn} for convolution blocks and LN \cite{ln} for transformer or MLP blocks by default. GELU \cite{gelu} is utilized as the activation function. We use CPE \cite{cpe} as the default positional embedding for transformer blocks, which is easy to transfer to larger input resolution.

We utilize the compound scaling \cite{efficientnet} to scale depth, width, and resolution simultaneously. The details of UniNet-B1 to B6 are listed in Table~\ref{tab:uninet}.

\begin{table}[h]
    \centering
    \caption{UniNet-B1 to B6 architectures. $\mathtt{c}$ and $\mathtt{r}$ are channels and repeats, respectively. S denotes stage. Input size is set to be divisible by 32 as the total down-sampling stride is 32.}
    \begin{tabular}{c|cc|cc|cc|cc|cc|cc}
    \toprule
    & \multicolumn{2}{c}{B1} & \multicolumn{2}{c}{B2} & \multicolumn{2}{c}{B3} & \multicolumn{2}{c}{B4} & \multicolumn{2}{c}{B5} & \multicolumn{2}{c}{B6} \\
    & $\mathtt{c}$ & $\mathtt{r}$ & $\mathtt{c}$ & $\mathtt{r}$ & $\mathtt{c}$ & $\mathtt{r}$ & $\mathtt{c}$ & $\mathtt{r}$ & $\mathtt{c}$ & $\mathtt{r}$ & $\mathtt{c}$ & $\mathtt{r}$\\
    \midrule
    S0 & 48 & 2 & 48 & 3 & 56 & 3 & 64 & 4 & 64 & 5 & 96 & 6 \\
    S1 & 80 & 4 & 80 & 6 & 96 & 7 & 112 & 9 & 112 & 10 & 160 & 12 \\
    S2 & 128 & 4 & 128 & 6 & 160 & 7 & 192 & 9 & 224 & 10 & 256 & 12 \\
    S3 & 128 & 4 & 128 & 6 & 160 & 7 & 192 & 9 & 224 & 10 & 256 & 12 \\
    S4 & 256 & 8 & 256 & 12 & 288 & 14 & 352 & 18 & 448 & 20 & 512 & 24 \\
    \midrule
    \makecell{Input \\ size} & \multicolumn{2}{c|}{224} & \multicolumn{2}{c|}{256} & \multicolumn{2}{c|}{288} & \multicolumn{2}{c|}{320} & \multicolumn{2}{c|}{384} & \multicolumn{2}{c}{448} \\
    
    \bottomrule
    \end{tabular}
    \label{tab:uninet}
\end{table}

\subsection{Inference Throughput}
\begin{figure}[h]
    \centering
    \vspace{-1em}
    \includegraphics[width=0.6\linewidth]{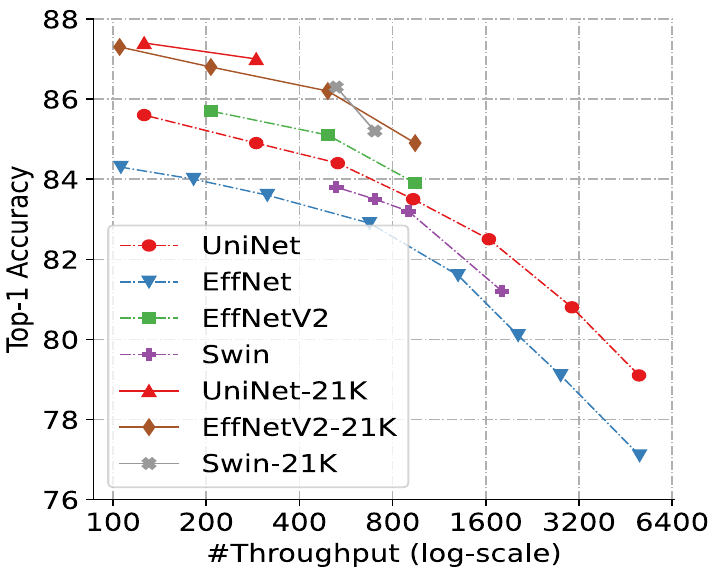}
    \caption{\textbf{Performance vs. inference throughput.} The throughput is measured on 80GB A100 GPU FP32 with batch size 128 using timm library\protect\footnotemark.}
    \label{fig:latency}
\end{figure}

We show the results in Figure~\ref{fig:latency}. The search of UniNet is based on optimizing FLOPs, and we believe directly optimizing latency could bring a better trade-off (like EffNetV2). Besides, we note that hardware speed heavily depends on software optimization. We believe that further optimization could bring the speed close to the theoretical optimum (FLOPs).

\footnotetext{\url{https://github.com/rwightman/pytorch-image-models/blob/master/benchmark.py}}

\subsection{Training Details}

\subsubsection{ImageNet-1K}
To train UniNet on ImageNet-1K, we follow the training receipt in DeiT \cite{deit}. We do not use repeated augmentation \cite{repeated} as it slows down the convergence process \cite{earlyconv}.
Following \cite{cait,effnetv2,coatnet,trainvit}, we apply higher drop path rate $d_r$ on large models, as shown in Table~\ref{tab:hp}. UniNet-B5 and UniNet-B6 are trained with $224 \times 224$ input size and finetuned on larger resolutions for training efficiency. The default training setting is in Table \ref{tab:traing_setting}. 
For finetuning, the weight decay rate is set to $10^{-8}$ \cite{vit}, the $\alpha$ in Mixup \cite{mixup} or CutMix \cite{cutmix} is set to 0.1. We list the detailed finetune setting in Table \ref{tab:finetune}.

\begin{table}[t]
    \centering
    \caption{Drop path rate for training UniNet models.}
    \begin{tabular}{c|c|c|c|c|c|c|c}
    \toprule
    Model & B0 & B1 & B2 & B3 & B4 & B5 & B6 \\
    \midrule
    $d_r$ & 0 & 0 & 0.1 & 0.1 & 0.3 & 0.5 & 0.6 \\
    \bottomrule
    \end{tabular}
    \label{tab:hp}
\end{table}

\begin{table}[t]
    \centering
    \caption{Training settings on ImageNet-1K.}
    \begin{tabular}{l|c}
        \toprule
        config & pretrain \\
        \midrule
        optimizer & AdamW \cite{adam} \\
        learning rate & 0.001  \\
        weight decay & 0.05  \\
        batch size & 1024  \\
        learning rate schedule & cosine decay \\
        warmup epochs & 5 \\
        training epochs & 300 \\
        augmentation & RandAug(9, 0.5) \cite{randaug} \\
        LabelSmooth \cite{inception} & 0.1 \\
        Mixup \cite{mixup} & 0.8 \\
        CutMix \cite{cutmix} & 1.0 \\
        drop path \cite{droppath} & Table \ref{tab:hp} \\
        \bottomrule
    \end{tabular}
    \label{tab:traing_setting}
\end{table}

\begin{table}[t]
    \centering
    \caption{Finetune settings on ImageNet-1K.}
    \begin{tabular}{l|c}
    \toprule
    config & value \\
    \midrule
    optimizer & AdamW \cite{adam} \\
    learning rate & 2.5e-6 \\
    weight decay & 1e-8 \\
    batch size & 512 \\
    learning rate schedule & cosine decay \\
    warmup epochs & 0 \\
    training epochs & 30 \\
    augmentation & RandAug(9, 0.5) \cite{randaug} \\
    LabelSmooth \cite{inception} & 0.1 \\
    Mixup \cite{mixup} & 0.1 \\
    CutMix \cite{cutmix} & 0.1 \\
    drop path \cite{droppath} & Table \ref{tab:hp} \\
    \bottomrule
    \end{tabular}
    \label{tab:finetune}
\end{table}

\begin{table}[t]
    \centering
    \caption{Training settings on ImageNet-21K.}
    \begin{tabular}{l|c}
    \toprule
    config & value \\
    \midrule
    optimizer & AdamW \cite{adam} \\
    learning rate & 0.0006 \\
    weight decay & 0.005 \\
    batch size & 2048 \\
    learning rate schedule & cosine decay \\
    warmup epochs & 5 \\
    training epochs & 90 \\
    augmentation & RandAug(9, 0.5) \cite{randaug} \\
    LabelSmooth \cite{inception} & 0.1 \\
    Mixup \cite{mixup} & 0.0 \\
    CutMix \cite{cutmix} & 0.0 \\
    drop path \cite{droppath} & 0.1 (B5), 0.4 (B6) \\
    \bottomrule
    \end{tabular}
    \label{tab:training_21k}
\end{table}

\subsubsection{ImageNet-21K}
To train UniNet on large-scale ImageNet-21K, we turn off the CutMix and Mixup augmentation. We utilize AdamW optimizer and train UniNet for 90 epochs. The detailed hyperparameters are in Table \ref{tab:training_21k}.

\subsubsection{ADE20K}
For transferring to ADE20K, we directly follow the task layer and most of the hyperparameters described in SETR-PUP \cite{setr}. The detailed hyperparameters are described in Table \ref{tab:ade20k}.

\begin{table}[h]
    \centering
    \caption{Training settings on ADE20K.}
    \begin{tabular}{l|c}
        \toprule
        config & value \\
        \midrule
        optimizer & Adam \cite{adam} \\
        learning rate & 0.001 \\
        weight decay & 0.05 \\
        batch size & 16 \\
        learning rate schedule & linear \\
        warmup steps & 1500 \\
        training steps & 160K \\
        input resolution & 512 $\times$ 512 \\
        \bottomrule
    \end{tabular}
    \label{tab:ade20k}
\end{table}

\subsubsection{Knowledge Distillation of UniNet-B0}
We utilize the vanilla Knowledge Distillation (KD) algorithm to train UniNet-B0 for comparison with other approaches. Formally, we consider a student model $\mathcal{S}$ with parameters $\theta_s$ for distilling the teacher. We denote the output logits from the student and the teacher as $z_s$ and $z_t$, respectively. The student is trained to minimize the cross-entropy (CE) loss between its predicted probability $p_s$ and the teacher's output probability $p_t$, where $p_s$ and $p_t$ are softened by a temperature ${\tau}$. We ignore the one-hot labels from ImageNet-1K for simplicity. The ${\tau}$ is set to 1.0 in our experiments.

\begin{align}
        & \min_{{\theta}_{s}} ~ {\tau}^2\mathrm{CE}(p_s, p_t), \label{eq:obj}\\
        \text{where} ~~ p_s=~&\mathrm{softmax}(z_s/\tau), \ p_t=\mathrm{softmax}(z_t/\tau). \nonumber
\end{align}

For a more fair comparison with previous approaches \cite{alphanet,ofa,levit}, we utilize the RegNetY-16G (83.6\%) \cite{regnet} as the teacher network.

\subsection{Neural Architecture Search (NAS) with Reinforcement Learning}
\begin{figure}[h]
    \centering
    \includegraphics[width=0.9\textwidth]{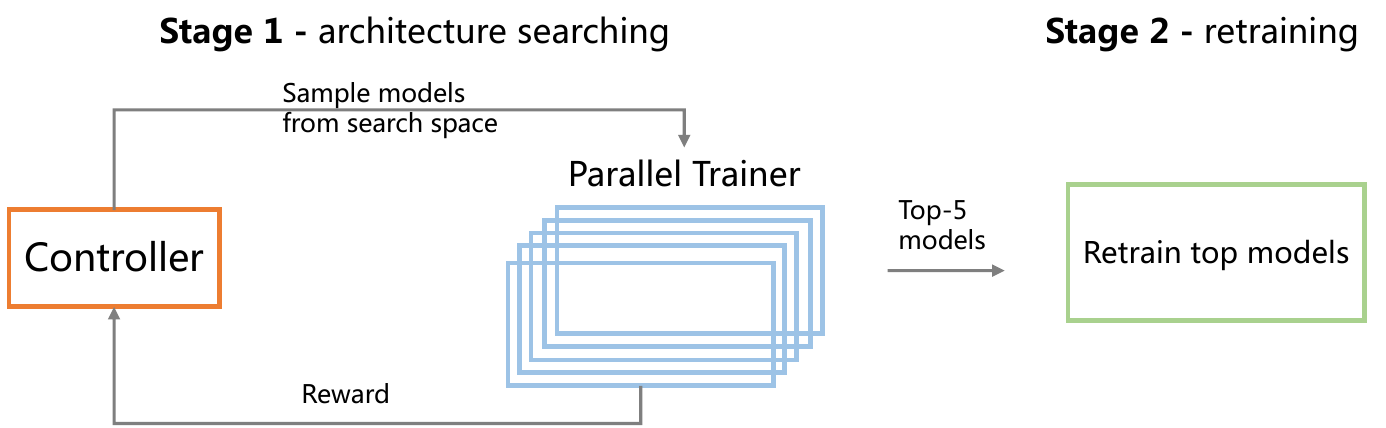}
    \caption{An overview of the NAS pipeline.}
    \label{fig:nas}
\end{figure}

We use the PPO \cite{ppo} algorithm to update the Reinforcement Learning (RL) agent. The RL agent is implemented with a two-layer LSTM with 100 hidden units at each layer. We use Adam \cite{adam} optimizer to train the RL agent with learning rate of 0.0005 and weight decay of 0. The weights of the controller are initialized uniformly between -0.1 and 0.1. We parallelly use about 80 GPUs to train the sampled architectures. We visualize the NAS pipeline in Figure \ref{fig:nas}. Note the other NAS algorithms such as one-shot-based \cite{ofa} or differentiable-based \cite{Liu2019DARTSDA} may also be applicable but are beyond the paper's scope.

\subsection{Visualization of MLP-Mixer}
In our searched UniNet, there is no MLP operator chosen. We find that the MLP-style operation breaks the spatial structure which is important for visual tasks \cite{islam2020much}, leading to sub-optimal performance when combined with other operators. We visualize the hidden features in Figure~\ref{fig:backbone}.

\begin{figure*}[t]
    \centering
    \includegraphics[width=0.9\textwidth]{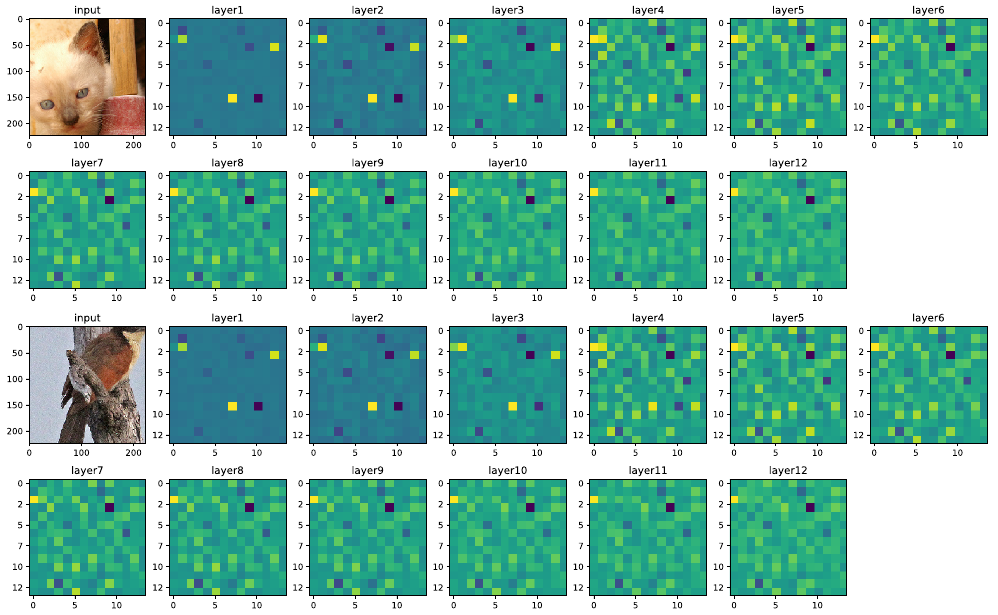}
    \caption{Hidden features of MLP-Mixer \cite{mixer}. Hidden features are reshaped back to 2D dimension.}
    \label{fig:backbone}
\end{figure*}

\clearpage
%
%
\bibliographystyle{splncs04}
\bibliography{egbib}
\end{document}